# Error Resilient Deep Compressive Sensing

*Thuong Nguyen Canh** and *Chien Trinh Van*#

*Department of Electronic and Electrical Engineering, Sungkyunkwan University, Korea
#Department of Electrical Engineering, Linkoping University, Sweden
*ngcthuong@skku.edu, #trinh.van.chien@liu.se

**ABSTRACT**

Compressive sensing (CS) is an emerging sampling technology that enables reconstructing signals from a subset of measurements and even corrupted measurements. Deep learning-based compressive sensing (DCS) has improved CS performance while maintaining a fast reconstruction but requires a training network for each measurement rate. Also, concerning the transmission scheme of measurement lost, DCS cannot recover the original signal. Thereby, it fails to maintain the error-resilient property. In this work, we proposed a robust deep reconstruction network to preserve the error-resilient property under the assumption of random measurement lost. Measurement lost layer is proposed to simulate the measurement lost in an end-to-end framework.

## 1. INTRODUCTION

Compressive sensing (CS) [1], an emerging sampling technique, is well-known for its unique functionality of simultaneous sampling, compression, and democracy (which enables error-resilient property [2, 3]). By capturing a sparse or compressible signal, $x \in \mathbb{R}^N$, with a random linear projection $\mathbf{\Phi} \in \mathbb{R}^{M \times N}$, at a smaller rate $M \ll N$ as

$$y = \mathbf{\Phi}x. \qquad (1)$$

with a random matrix $\mathbf{\Phi}$, each $M$ measurements in $y \in \mathbb{R}^M$ carries same amount of signal information thus equally important. As a result, it is possible to recover the original signal in case of missing a few measurements, thereby, enables the error-resilient property [3]. On the other hand, due to the fully random matrix and utilizing only sparse assumption, CS requires huge storage and computation for sampling (reduced by block based compressive sensing – BCS [4], separable sampling with Kronecker CS – KCS [5], sparse and binary sampling matrix, structure sampling matrix, etc.), suffers low efficient sensing (improved by utilizing other signal priors like low-frequency prior in multi-scale sampling [6, 7], structure sparse prior in weighted sampling [8, 9], etc.), and high complexity of reconstruction (improved by fast recovery algorithm [10]).

Recently, deep learning (DL) has been applied in compressive sensing to improve sampling efficiency with a non-iterative reconstruction [11, 12]. By removing the bias $\boldsymbol{b}$ and the activation $\boldsymbol{\sigma}$, the fully connected layer

$$y = \sigma(Wx + b), \qquad (2)$$

becomes the sampling in CS. Thus, research on Deep Learning based Compressive Sensing (DCS) has moved from learning to reconstruction from CS measurement [11, 12] to joint learn the sampling and reconstruction simultaneously [13, 14, 15]. Unlike the conventional CS in which one reconstruction can recover signals from different CS sampling scheme, the DCS based reconstruction is designed for the learned sampling matrix, thus produces higher quality. Additionally, DCS also reduces the computation complexity of sampling by following the block-based sampling via convolution [14], separable sampling KCS [15]. In general, researchers designed DCS to mimic the conventional single scale CS sampling with [14, 15] (i) sampling, (ii) initial reconstruction, and (iii) enhanced reconstruction. Multi-scale DCS was developed in [16] with multiple phase training.

However, DCS suffers from a practical problem of using many training networks [17]. If there are $n$ different subrates, $n$ corresponding networks are required to train. Therefore, Shi et al. proposed to initially reconstructs CS measurement at a different subrate then enhance reconstruction with a residual network and a total variation loss function. By doing so, multiple subrates share a same residual enhanced network. Unfortunately, their solution is suitable for a discrete subrate scenario only. Each subrate requires a different sampling, and initial reconstruction network.

On the other hand, the corrupted measurements in transmission can be considered as a reduced measurement rate. As the rate of missed measurement is continuous, thereby, the problem of continuous measurement rate. Shi's method in [17] still requires tremendous sampling and initial reconstruction layers which is impractical to maintain the error resilient applications.

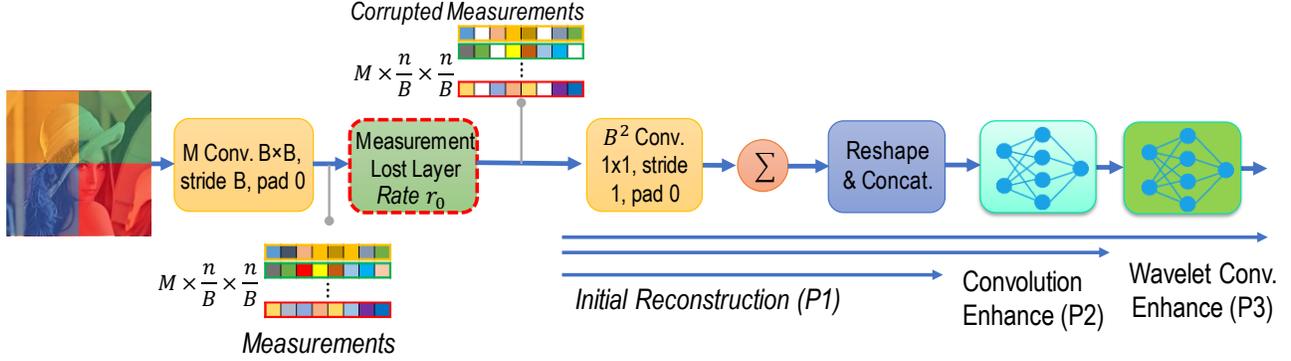

Fig. 1. Proposed multi-rate error resilient networks. The proposed method reconstructs image measurement loss at rate $r_1$. Note that, unlike Dropout, measurement lost layer is remain active during the testing phase.

In this work, we first proposed a measurement lost layer to mimic the missing measurements due to transmission and discuss its connection with the dropout layer [18] in Section 2. To further improve the quality of reconstruction especially in the case of measurement lost, we conduct exhaustive experiments with results are presented in Section.

## 2. PROPOSED MEASUREMENT LOST LAYER

### 2.1. Measurement Lost Layer

Unlike other degraded measurement scenario, the measurement lost is the scenario that some measurements are not received at the decoder side due the poor transmission. To model this measurement lost at the decoder side, we zero out the lost measurement (i.e., the value of lost/corrupted measurements are set to zero). By doing so, we introduce a Measurement Loss Layer (MLL), that mimic the measurement lost by

$$y = \mathcal{M} \odot x$$

where, $\mathcal{M} \in \mathbb{B}^{w \times h \times c}$ denotes a binary selection matrix with the identical size as the input signal $x \in \mathbb{R}^{w \times h \times c}$. The value of matrix $\mathcal{M}$ is generated follow the Bernoulli distribution at the ratio $r$ to mimic the random measurement lost. It can be implemented with any binary distribution toward the real lost model or channel burst error. In this work, $r = 0.1$ means that 10% of measurements are randomly lost.

### 2.2. Connection with Dropout Layer

One can be noticed that the proposed layer shares some similarity with the well-known dropout layer. While dropout layer follows the Bernoulli distribution. Additionally, dropout rate is set to a fixed ratio which is often as large as 0.5 to reduces the features size and increase the robustness of the network. Especially, dropout is only turn on during the training phase but not testing phase. That is, features are not dropped out during the inference process.

On the other hand, our proposed Measurement Lost layer is set with a small drop ratio and varied from 0~0.3, where 0 means no measurement lost. It should be noted that our MML layers are being active in both the training and testing phase. Additionally, the location of lost measurements of MML layers should follow the distribution of the simulated channel which mimic the transmission channel model. In conclusion, our MML is a generalized version of dropout layer, which aims not to increase the robustness of the network but to simulate the lost measurement scenario due to the bad channel condition.

### 2.3. Implementation Network

To verify the effective of our measurement lost layer, we incorporate MML to our previous work on the multi-scale deep compressive sensing network named Wavelet-based Deep Compressive Sensing (WDCS) [16]. In this prior work, we utilized the wavelet transform to decomposed signal into four subbands and sampling across all subbands. The network is trained in three stages with the initial reconstruction (P1) with a simple network, the enhancement with convolution (P2) and the final enhancement with the multi-level wavelet convolution (P3).

As illustrated in Fig. 1, the proposed measurement loss layer at a given rate is added after the conventional multi-scale sampling layer of WDCS. The other layers are identical to the original implementation of WDCS. We also use three phase training as in WDCS. Additionally, we implement two variation of our measurement lost layer.

In the first scenario, the measurement lost rate is fixed during the training and testing thus denoted as Fix-WDCS. In the second scenario, we vary the drop measurement rate during the training process as well as the testing phase, thereby, named Vary-WDCS. The varying rate is decided as 0 ~ 0.1, and 0.1 ~ 0.2 to make it equivalent to the Fix-WDCS at drop rate 0.1, 0.2, and 0.3.

Ones should be noted that, MLL in Fix-WDCS is more like Dropout layer while it is more like adaptive Dropout in Vary-WDCS.

Table I. Performance for deep compressive sensing networks under various measurement drop rates for test image Set5.

| Set5 | Train Drop Rate | Sub-rate | Test Drop Rate | | | | | | | | | | | |
|---|---|---|---|---|---|---|---|---|---|---|---|---|---|---|
| | | | 0 | | | 0.1 | | | 0.2 | | | 0.3 | | |
| | | | P1 | P2 | P3 | P1 | P2 | P3 | P1 | P2 | P3 | P1 | P2 | P3 |
| WDCS [16] | 0 | 0.1 | 30.60 | 32.44 | 33.39 | 19.49 | 18.01 | 18.08 | 16.23 | 14.67 | 14.54 | 14.28 | 12.69 | 12.45 |
| | | 0.2 | 34.06 | 35.99 | 35.56 | 19.43 | 18.81 | 19.76 | 16.10 | 15.35 | 15.66 | 13.71 | 12.96 | 13.00 |
| Fixed-WDCS | 0.1 | 0.1 | 30.10 | 25.76 | 24.86 | 23.70 | **29.07** | **30.49** | 23.71 | 23.12 | 23.57 | **21.95** | **18.26** | **18.02** |
| | | 0.2 | 33.01 | 26.34 | 27.14 | 28.07 | **31.89** | **32.87** | 25.54 | 23.59 | 23.91 | 23.64 | 17.98 | **19.00** |
| | 0.2 | 0.1 | 28.64 | 20.58 | - | 25.99 | 24.38 | - | **24.01** | **28.23** | - | 22.39 | 21.91 | - |
| | | 0.2 | 31.31 | 20.84 | - | 27.99 | 25.17 | - | **25.91** | **29.74** | - | **23.97** | 22.73 | - |
| Vary-WDCS | 0.1 | 0.1 | 30.56 | 31.14 | - | **26.11** | 28.03 | - | 23.52 | 21.75 | - | 21.64 | 17.4 | - |
| | | 0.2 | 33.61 | 34.88 | - | **28.17** | 29.69 | - | 25.39 | 26.41 | - | 23.41 | **24.02** | - |
| | 0.2 | 0.1 | 30.86 | 32.51 | 33.26 | 20.38 | 18.62 | 18.43 | 16.79 | 14.85 | 14.97 | 14.56 | 12.61 | 12.56 |
| | | 0.2 | **34.11** | **36.01** | **36.58** | 20.27 | 19.75 | 19.75 | 16.08 | 15.91 | 15.91 | 14.06 | 13.43 | 13.76 |

*P1, P2, and P3 are three steps of network WDCS.*

## 3. EXPERIMENTAL RESULTS

### 3.1. Experimental Setting

**Training data**. In this section, we first describe the experimental condition, then show the reconstruction performance. This work uses DIV2K [16] for training with 64×500 patches of size 256×256 and implement with MatConvNet [18], tested with 6 test images of Set5. For the DL-based compressive sensing network, we use multi-scale wavelet deep compressive sensing (WDCS) as the baseline (trained at the drop rate 0). Our proposed method Fix-WDCS and Vary-WDCS are trained under various drop rate scenarios of 0.1 and 0.2. Test drop rate are 0, 0.1, 0.2 and 0.3. Similar to the previous DCS network. We use the mean square error as the loss function, Adam as the optimizer.

### 3.2. Simulated Results

The experimental results are shown in Table. 1 with a different combination of training and testing drop rate. Firstly, it is easy to observe that the network has a strong connection with its learned measurement as WDCS reduces significantly performance event at the low-drop rate of 0.1. In addition, a more complex network (at stage P3), performance drop increase.

On the other hand, with MLL at a fixed drop rate, Fix-WDCS reduces quality in the case of no measurement drop around 1dB. Fix-WDCS, however, significant increase the reconstruction quality in case of measurement drop. It improves 12.30 dB at test drop 0.1 and train drop 0.1 over WDCS. Even at high drop rate 0.3, Fix-WDCS at 0.1 train drop rate still 6dB better that the WDCS. Our Vary-WDCS also produce better reconstruction quality than WDCS but much less than Fixed-WDCS in the presence of measurement loss at 0.4 ~ 0.5 dB improvement. Interestingly, Vary-WDCS significantly boost reconstruction quality under the case of no measurement loss at a high subrate. An improvement of 1.02 dB is present at Vary-DWCS train drop rate 0.2 over WDCS. Therefore, with a careful designed varying drop-rate (or adaptive drop-rate), it is possible to increase the robustness of the network and resulted in better reconstruction performance.

In addition, as we observe that the more complex network is, the more loss in performance under higher measurement drop rate. This is due to the binding of both reconstruction and sampling network as they are jointly learned. While this relationship increases the reconstruction quality under perfect transmission scenario, a single measurement lost in the transmission could results in heavily degraded in the reconstruction quality. Therefore, in future work, we would like to utilize a multiple shadow networks for initial reconstruction image reconstruction, and a complex network will be utilized to enhance the reconstruction performance.

## 4. CONCLUSION

In this paper, we proposed a measurement lost layer to preserves error resilient capability of deep learning based compressive sensing. The proposed layer can be fixed or varying drop rate. While fixed drop rate layer significantly increases the reconstruction quality at the present of measurement loss, the vary drop rate layer boost reconstruction performance at a higher subrate. Currently, the simulation results are limited to Bernoulli random. In the future work, other distribution of loss measurements can be modelled to match the channel modelling in communication.